\documentclass[12pt]{article}
\usepackage[utf8]{inputenc}

\usepackage{graphicx}

\usepackage{algorithmic}
\usepackage{verbatim}
\usepackage{mathtools}
\usepackage{graphicx}
\usepackage{tikz}
\usepackage{stmaryrd}
\usepackage{booktabs}
\usepackage{amsfonts}
\usepackage{amssymb}
\usepackage{amsmath}
\usetikzlibrary{positioning}
\usepackage{caption}
\usepackage{subcaption}
\usepackage{diagbox}
\usepackage{blkarray}
\usepackage{multirow}
\usepackage{hyperref}
\usepackage{caption}
\usepackage[autostyle]{csquotes}  
\usepackage{float}

\DeclareMathOperator*{\argmax}{arg\,max}

\title{Natural Policy Gradients In Reinforcement Learning Explained}
\author{W.J.A. van Heeswijk}

\begin{document}
\begin{titlepage}
\maketitle
\end{titlepage}

\begin{abstract}
Traditional policy gradient methods are fundamentally flawed. Natural gradients converge quicker and better, forming the foundation of contemporary Reinforcement Learning such as Trust Region Policy Optimization (TRPO) and Proximal Policy Optimization (PPO). This lecture note aims to clarify the intuition behind natural policy gradients, focusing on the thought process and the key mathematical constructs. 
\end{abstract}

\section{Introduction}
Policy gradient algorithms are at the root of modern Reinforcement Learning. The idea is that, by simply following the gradient (i.e., vector of partial derivatives) of the objective function, we ultimately end up at an optimum. It is a clever way to (i) directly optimize policies (rather than learn indirect value functions) and (ii) let the reward function guide the search. However, policy gradients have fundamental flaws. This article explains the concept of natural gradients, exposing the shortcomings of traditional gradients and how to remedy them.

Although natural gradients have been surpassed in popularity by algorithms such as TRPO and PPO, a grasp on their fundamentals is essential to understand these contemporary RL algorithms. Natural policy gradients deploy different way of thinking, which is not always clear from just observing the loss function.
Nonetheless, a complete discussion of natural gradients is rather technical and requires many lengthy derivations. To keep this article (somewhat) compact, I focus mainly on the reasoning and intuition, providing external references to more in-depth derivations. Furthermore, a solid understanding of traditional (vanilla) policy gradients and the REINFORCE algorithm is assumed.

\section{The problems with first-order policy gradients}
In traditional policy gradient methods, the gradient $\nabla$ only gives the direction of the weight update. It does not tell how far to step into that direction. Derivatives are defined on an infinitesimal interval, meaning the gradient is valid only locally, and may be completely different at another part of the function.

Because of this, we iterate between sampling (with the current policy) and updating (based on sampled data). Each new policy rollout allows re-computing the gradient and update the policy weights $\theta$.

Behavior is controlled with the step size $\alpha$. This gives rise to the following well-known policy gradient update function:

\begin{figure}[H]
\centering
\begin{align}\notag
\theta \mapsfrom \theta + \alpha \nabla_\theta J(\theta)
\end{align}
\caption*{\textit{Traditional policy gradient update function, updating policy weights $\theta$ based on objective function gradient $\nabla_\theta J(\theta)$ and step size $\alpha$}}
\end{figure}

Two common problems may arise during updates:

\begin{itemize}
\item	Overshooting: The update misses the reward peak and lands in a suboptimal policy region.
\item	Undershooting: Taking needlessly small steps in the gradient direction causes slow convergence
\end{itemize}

In supervised learning problems, overshooting is not too big of a deal, as data is fixed. If we overshoot, we can correct next epoch. However, if an RL update results in a poor policy, future sample batches may not provide much meaningful information. Somewhat dramatically: we may never recover from a single poor update. A very small learning rate might remedy the problem, but results in slow convergence.

\begin{figure}[H]
\includegraphics[width=\textwidth]{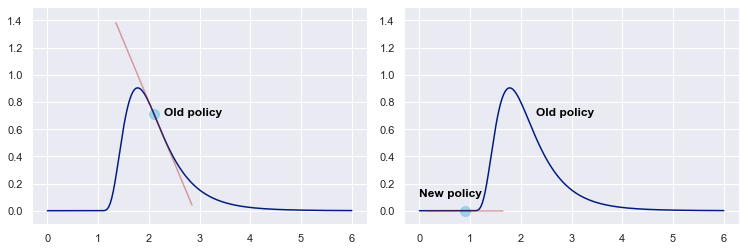}
\caption{Example of overshooting. If the step taken into the gradient direction is too large (left), the update might miss the reward peak and land in a suboptimal region with low gradients (right).}
\end{figure}
 
After the update, we landed on a flat, suboptimal region. The low gradients stir only small weight updates, and it will take many iterations to escape again.

An interesting observation here is that we performed a large update when we should have performed a cautious one, and vice versa. As we will see later, natural gradients do the opposite.

Let’s perform a thought experiment. To take appropriately-sized weight updates, we might decide to put a cap on the parameter changes. Suppose we define a maximum distance in parameter space as a constraint. We could define this problem as follows:

\begin{figure}[H]
\centering
\begin{align}
	{\Delta \theta^*} = \argmax_{\lVert \Delta \theta \rVert \leq \epsilon} J(\theta+\Delta \theta)\notag
\end{align}
\caption*{\textit{A weight update scheme that caps the Euclidean distance between old and updated parameters.}}
\end{figure}
 
where $\lVert \Delta \theta \rVert$ represents the Euclidian distance between parameters before and after the update.

It sounds reasonable, as it should avoid overshooting, while also not needlessly restricting the update size. Unfortunately, it does not work as you might expect. For instance, suppose our policy is a Gaussian control parameterized by $\theta_1=\mu$ and $\theta_2=\sigma$, and we set a cap $\epsilon=1$. Both updates in the figure below satisfy the constraint!

\begin{figure}[H]
\includegraphics[width=\textwidth]{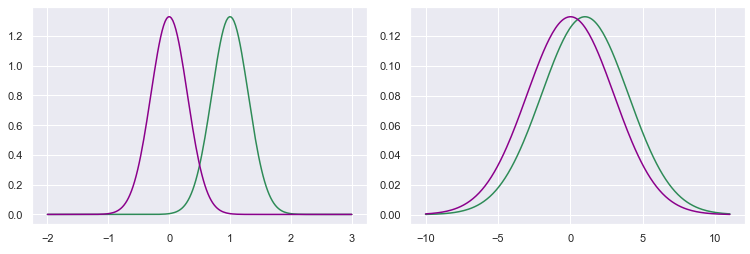}
\caption{Comparison of normal distribution pairs. The left has $\mu_1=0$, $\mu_2=1$ and $\sigma_1=\sigma_2=0.3$. The right has $\mu_1=0$, $\mu_2=1$ and $\sigma_1=\sigma_2=3.0$. Although the Euclidean distance between both pairs is 1, it is obvious the pair on the right is much more similar than the one on the left.}
\end{figure}
 
In both cases, the Euclidean distance is 1: $\sqrt{(1–0)^2+(0.3–0.3)^2}$ and $\sqrt{(1–0)^2+(3–3)^2}$. However, the effect on the distribution (i.e., the stochastic policy) is completely different. 

The problem is that capping the parameter space does not effectively cap the statistical manifold that we operate on. Remind that policies are probability distributions, and altering the probabilities alters the expected rewards. This is the manifold we optimize over and would like to control.

 \begin{quotation}
\textit{\noindent Note: I like to think of a statistical manifold as a 'family’ of distributions. For instance, the collective family of normal distributions (parameterized by $\mu$ and $\sigma$) constitutes a manifold. Another example would be to consider a neural network as a distribution over output values. Altering a stochastic policy can be viewed as moving over the manifold (occupied by the parameters $\theta$) over which the objective function is defined.}
\end{quotation}

The parameter cap only works as intended if the statistical manifold is linear, which is rarely the case. To prevent the policy itself from changing too much during an update, we must consider how sensitive the distribution is to parameter changes. Traditional policy gradient algorithms do not take this curvature into account. To do that, we need to move into the realm of second-order derivatives, which is exactly what natural policy gradients do.

\section{Capping the difference between policies}
We established that the difference between distributions (i.e., policies parameterized by $\theta$) is of interest, rather than the difference between parameters $\theta$ and $\theta_\text{old}$ itself. Fortunately, a variety of distances exist to calculate the difference between two probability distributions. This article will use the KL divergence, which is most common in literature. Technically it is not a measure (as it is not symmetric), but can be thought of as such (being approximately symmetric for small discrepancies):

\begin{figure}[H]
\centering
\begin{align} 
	\mathcal{D}_\text{KL}(\pi_{\theta} \parallel \pi_{\theta+\Delta\theta}) = \sum_{x\in\mathcal{X}} \pi_{\theta}(x) \log\left(\frac{\pi_{\theta}(x)}{ \pi_{\theta+\Delta\theta}(x)}\right) \notag
\end{align}
\caption*{\textit{Kullback-Leibner divergence (also ‘relative entropy’) between policies $\pi$ and $\pi_\text{old}$. It describes the distance between two probability distributions.}}
\end{figure}

 \begin{quotation}
\textit{\noindent Note: in the normal distribution example shown before, KL divergences are 0.81661 and 0.023481, respectively.}
\end{quotation}

At this point, it is good to introduce the link between KL divergence and the Fisher information matrix (we will see why later). The Fisher information matrix is the Riemannian metric describing the curvature of a statistical manifold, i.e., the sensitivity of the manifold to marginal parameter changes. The matrix may be viewed as a correction to the distance that accounts for the curvature--think of measuring distances on a globe rather than on a flat earth.

If we define KL divergence locally, i.e., $\Delta\theta=0$, it turns out both are equivalent. Under this condition, the zeroth and first derivatives become 0 and can be stricken. The matrix of second derivatives is represented by the Hessian matrix, which in this case is equivalent to the Fisher information matrix:

\begin{figure}[H]
\centering
\begin{align}
F(\theta) = \nabla_\theta^2 \mathcal{D}_{KL}(\pi_{\theta}(x) || \pi_{\theta + \Delta \theta}(x))|_{\Delta \theta = 0} \notag
\end{align}
\caption*{\textit{Locally, KL divergence is equivalent to the Fisher matrix. This result is helpful for practical implementations.}}
\end{figure}
 
This result will prove crucial for the practical implementation, but let’s put a pin in that for now.

 \begin{quotation}
\textit{\noindent Note: If the Fisher matrix is an identity matrix ($F(\theta)=I$), the distance over the manifold is simply the Euclidean distance. In that case, traditional- and natural policy gradients are equivalent. In practice, this is rare though.}
\end{quotation}

Similar to before, we place a constraint on the allowed change of an update. This time, however, we impose it on the KL divergence of the policy, rather than on the Euclidean distance of the parameter space. The adjusted problem looks as follows:

\begin{figure}[H]
\centering
\begin{align}
\Delta \theta^* = \argmax_{ \mathcal{D}_\text{KL}(\pi_\theta \parallel \pi_{\theta+\Delta \theta}) \leq \epsilon} J(\theta+\Delta \theta) \notag
\end{align}
\caption*{\textit{A weight update scheme that caps the KL divergence between old and updated policy. Note that this scheme considers the discrepancy between distributions, not parameters.}}
\end{figure}
 
By solving this expression, we ensure that we perform a large update in parameter space, while ensuring the policy itself does not change too much. However, computing the KL divergence requires evaluating all state-action pairs, so we need some simplifications to work with realistic RL problems.

\section{Lagrangian relaxation and Taylor expansion}
For the upcoming section, an excellent and detailed derivation can be found in the lecture slides from Carnegie Mellon (by Katerina Fragkiadaki, see end of paper). To preserve focus on the intuition, I only highlight the most salient outcomes.

We are going to derive the solution for the problem. First, we use Lagrangian relaxation to transform the divergence constraint into a penalty, yielding an expression that is easier to solve:

\begin{figure}[H]
\centering
\begin{align}
\Delta\theta^* = \argmax_{\Delta\theta} J(\theta+\Delta\theta)-\lambda( \mathcal{D}_\text{KL}(\pi_\theta \parallel \pi_{\theta+\Delta \theta})-\epsilon)\notag
\end{align}
\caption*{\textit{ By performing Langrangian relaxation, we obtain an expression that penalizes large policy changes rather than constraining them. This expression is easier to solve.}}
\end{figure}
 
Given that typical RL problems are way too large to compute the divergence $\mathcal{D}_{\text{KL}}$ for all states and actions, we must resort to approximation methods. With Taylor expansion--approximating a function in terms of its derivatives--we can approach the KL divergence based on sample trajectories obtained with the rollout policy $\pi_\theta$.

The Taylor expansion for the above-mentioned Lagrangian relaxation looks as follows (for notational convenience, consider $\theta=\theta_\text{old}+\Delta\theta$):

\begin{figure}[H]
\centering
\begin{align}
	\Delta\theta^* \approx \argmax_{\Delta\theta} J(\theta_\text{old})+ \nabla_{\theta} J(\theta)|_{\theta=\theta_\text{old}}\cdot \Delta \theta \notag\\
	-\frac{1}{2}\lambda\left(\Delta \theta^\top \nabla_{\theta}^2 \mathcal{D}_\text{KL}(\pi_{\theta_\text{old}} \parallel \pi_{\theta})|_{\theta=\theta_\text{old}}\Delta \theta\right)+\lambda\epsilon \notag
\end{align}
\caption*{\textit{ Taylor expansion to approximate the optimal weight update scheme. The expansion takes the first-order expansion of the loss and the second-order expansion of the KL divergence.}}
\end{figure}
 
In short, the loss term $J(\theta)$ is approximated with the first-order Taylor expansion (i.e., the gradient w.r.t. $\theta$), similar to traditional policy gradients (essentially, local linearization). The KL divergence is approximated with a second-order Taylor expansion. When locally approximating KL divergence (i.e., $\Delta\theta = 0$), the zeroth and first-order differences evaluate to 0, such that we can strike them. The second-order derivatives are of interest here. 
To make the expression a bit less intimidating, we can (i) replace the second-order KL derivative with the Fisher information matrix and (ii) strike all terms that do not depend on $\Delta\theta$. This leaves us with a slightly friendlier expression: 

\begin{figure}[H]
\centering
\begin{align}
	\Delta\theta^* \approx \argmax_{\Delta\theta}  \nabla_{\theta} J(\theta)|_{\theta=\theta_\text{old}}\cdot \Delta \theta-\frac{1}{2}\lambda\left(\Delta \theta^\top  F(\theta_\text{old})\Delta \theta\right) \notag
\end{align}
\caption*{\textit{Simplified Taylor expansion of the weight update scheme, substituting in the Fisher matrix and striking terms that do not depend on $\Delta\theta$}}
\end{figure}

Notational compactness aside, why substitute the second-order derivative with the Fisher matrix? Well, it turns out the equivalence is very convenient. The Hessian matrix is a $|\theta| \cdot |\theta|$ matrix, with each element being a second derivative. Full computation may be quite cumbersome. For the Fisher matrix we have an alternative expression however, which is the outer product of the gradients. As we already need these values for traditional policy gradients anyway, there is substantially less computational overhead:

\begin{figure}[H]
\centering
\begin{align}
	F(\theta)=\mathbb{E}_\theta \left[\nabla_{\theta} \log \pi_{\theta}(x) \nabla_{\theta} \log \pi_{\theta}(x)^\top \right] \notag
\end{align}
\caption*{\textit{The Fisher information matrix can be represented as the outer product of policy gradients. The expression is locally equivalent to the Hessian matrix, but computationally more efficient to generate.}}
\end{figure}
 
Thus, if we can compute the gradients as we are used to, we have all the information necessary to perform the weight update. Also note the expectation implies we can use samples.

This is quite some information to take in, so let’s briefly recap what we accomplished so far:

\begin{itemize}
\item	To prevent the policy drifting too far away, we placed a constraint on the KL divergence between new and old policy. 
\item	Using Lagrangian relaxation, we transformed the constraint into a penalty, giving us a single (unconstrained) expression to work with.
\item	As we cannot directly compute KL divergence based on samples, we used Taylor expansion as an approximation for the weight update scheme. 
\item	For small parameter changes, KL divergence is approximated using the Fisher information matrix, for which we have a readily available expression.
\item	The entire approximation is a local result, assuming $\theta=\theta_\text{old}$. As such, the entire rationale is only valid for small policy changes.
\end{itemize}

Now, let’s see how we can solve this problem.

\section{Solving the KL-constrained problem}
Time to circle back to our Taylor expansion of the Lagrangian relaxation. How do we solve this expression, i.e., find the optimal weight update $\Delta\theta$?

\begin{figure}[H]
\centering
\begin{align}
	\Delta\theta^* \approx \argmax_{\Delta\theta}  \nabla_{\theta} J(\theta)|_{\theta=\theta_\text{old}}\cdot \Delta \theta-\frac{1}{2}\lambda\left(\Delta \theta^\top  F(\theta_\text{old})\Delta \theta\right) \notag
\end{align}
\caption*{\textit{The simplified Taylor expansion of the weight update scheme can be resolved using the Lagrangian method}}
\end{figure}

Well, we can find the desired update by setting the gradient w.r.t. $\Delta\theta$ to zero (Langrangian method). Solving the expression (converted to a minimization problem now, and assuming $\theta=\theta_\text{old}$) yields:

\begin{figure}[H]
\centering
\begin{align}
	0 &= \frac{\partial}{\partial \Delta\theta} \left( J(\theta) + \nabla_\theta J(\theta) \Delta\theta + \frac{1}{2} \lambda \, \Delta\theta^\top F(\theta) \Delta\theta \right) \notag\\
	&= \nabla_\theta J(\theta) + \lambda \, F(\theta) \Delta\theta   \notag
\end{align}
\caption*{\textit{ Solving the relaxed Taylor expansion by setting the derivative w.r.t. $\Delta\theta$ to 0}}
\end{figure}
 
The solution can be rearranged to find the weight update $\Delta\theta$:

\begin{figure}[H]
\centering
\begin{align}
	\lambda \, F(\theta) \Delta\theta &= -\nabla_\theta J(\theta) \notag \\
	\Delta\theta &= -\frac{1}{\lambda} F(\theta)^{-1} \nabla_\theta J(\theta) \notag
\end{align}
\caption*{\textit{Re-arranging the solution allows to express the optimal weight update}}
\end{figure}
 
Note that $\frac{-1}{\lambda}$ is a constant that can be absorbed into the learning rate $\alpha$. In fact, $\alpha$ can be derived analytically. From the initial constraint, we know that the KL-divergence should be at most $\epsilon$. With a fixed learning rate $\alpha$, we would be unable to guarantee that $\alpha F(\theta)^-1 \nabla_\theta J(\theta)\leq\epsilon$. Algebraically, we can thus infer a dynamic learning rate $\alpha$ that ensures (again, by approximation) the size of the update equals $\epsilon$. Abiding by this constraint yields the following learning rate:

\begin{figure}[H]
\centering
\begin{align}
	\alpha=\sqrt{\frac{2\epsilon}{\nabla J(\theta)^\top F(\theta)^{-1}\nabla J(\theta)}} \notag
\end{align}
\caption*{\textit{ The dynamic learning rate $\alpha$ ensures that the KL-divergence of the weight update (by approximation) does not exceed the divergence threshold $\epsilon$}}
\end{figure}

Finally, from the re-arrangement, we extract the \textit{natural} policy gradient, which is the gradient corrected for the curvature of the manifold:

\begin{figure}[H]
\centering
\begin{align}
	\tilde{\nabla} J(\theta)= F(\theta)^{-1} \nabla J(\theta) \notag
\end{align}
\caption*{\textit{ The natural policy gradient w.r.t. the objective function is the standard gradient multiplied with the inverse Fisher matrix, accounting for the curvature of the Riemannian space.}}
\end{figure}
 
This natural gradient gives -- within the distant constraint -- the steepest descent direction in the Riemannian space, rather than in the traditionally assumed Euclidean space. Note that, compared to traditional policy gradients, the only distinction is the multiplication with the inverse Fisher matrix! In fact, if the Fisher matrix is an identity matrix--which in practice it rarely is--traditional- and natural policy gradients are equivalent.

The final weight update scheme looks as follows:

\begin{figure}[H]
\centering
\begin{align}
\Delta \theta=  \sqrt{\frac{2\epsilon}{\nabla J(\theta)^\top F(\theta)^{-1}\nabla J(\theta)}} \tilde{\nabla} J(\theta) \notag
\end{align}
\caption*{\textit{ Weight update scheme for the natural policy gradient. The dynamic learning rate ensures that each update equally alters the distribution.}}
\end{figure}
 
The power of this scheme is that it always changes the policy by the same magnitude, regardless of the distribution’s representation.

The end result differs from traditional policy gradients in two ways:

\begin{itemize}
\item	The gradient is ‘corrected’ by the inverse Fisher matrix, taking into account the sensitivity of the policy to local changes. As the matrix is inverted, updates tend to be cautious at steep slopes (high sensitivity) and larger at flat surfaces (low sensitivity). Traditional gradient methods (erroneously) assume Euclidian distances between updates.
\item	The update weight/step size $\alpha$ has an dynamic expression that adapts to the gradients and local sensitivity, ensuring a policy change of magnitude $\epsilon$ regardless of parameterization. In traditional methods, $\alpha$ is a tunable parameter suspect to misfitting, often set at some standard value like 0.1 or 0.01.
\end{itemize}

Despite the considerably different mechanisms under the hood, at surface level traditional- and natural policy gradient methods are surprisingly similar.

The full outline of the natural policy gradient algorithm is summarized below. Note that in practice, we always use sample estimates for gradients and the Fisher matrix.

\begin{figure}[h!]
\includegraphics[width=\textwidth]{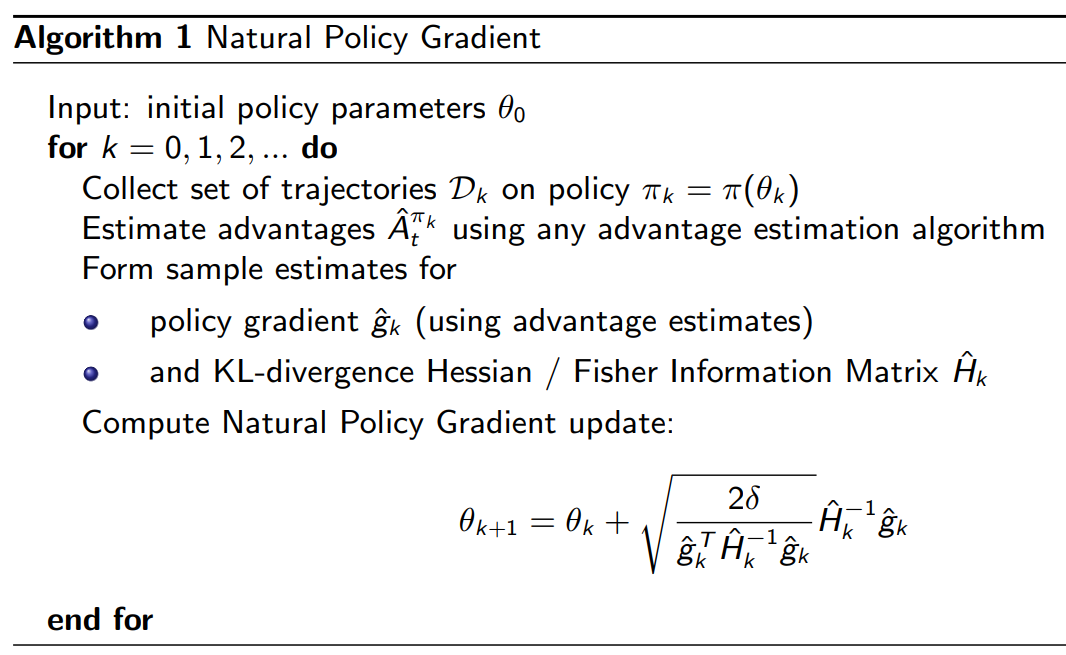}
\caption{Natural policy gradient algorithm, from Berkeley’s Deep RL course by Joshua Achiam}
\end{figure}
 
\section{Problems with natural gradients}
Natural gradients overcome fundamental flaws of traditional methods, taking into account how the manifold--over which the objective function is defined -- changes with parameter updates. Concretely, natural gradients allow to escape for plateaus and cautiously approach reward peaks. Theoretically, natural policy gradient should therefore converge better and faster than their traditional counterparts.

In their purest form, natural gradient algorithms are often not practical though. There are a number of reasons for this.

First, the Taylor expansion offers a local approximation up to the second order. Due to this, the estimated Hessian may not be positive definite. In practice, natural gradient methods are numerically brittle and do not always yield stable outcomes. The abundance of mathematical derivations may look convincing, but the Taylor expansion, sample approximations and strictly local validity (assuming $\theta = \theta_\text{old}$) substantially impact real-world performance.

Second, the Fisher information matrix occupies an $|\theta|\cdot \theta|$ space. Consider a neural network with 100,000 parameters, and you can imagine a 10 billion matrix on your laptop will not fly. Additionally, computing a matrix inverse is an operation of $\mathcal{O}(N^3)$ complexity, which is rather tedious. Thus, for deep RL methods, natural policy gradients typically exceed both memory- and computational limits.

Finally, we are used to work with sophisticated first-order stochastic gradient optimizers -- such as ADAM, which also takes into account second-order effects--that provide excellent results on a wide range of problems. Second-order optimization methods (i.e., natural gradient algorithms) do not take advantage of these optimizers.

Methods such as conjugate gradients and Kronecker-factored approximation curvature (K-FAC) may (partially) address the above-mentioned problems. In practice, methods such as Trust Region Policy Optimization (TRPO) and especially Proximal Policy Optimization (PPO) have surpassed natural gradients in popularity, although being rooted in the same mathematical foundation.

\section{Closing words}
When contrasting natural policy gradients to traditional ones, the difference looks fairly limited. In the end, we only added an inverted Fisher matrix--factoring in local sensitivity--to the gradient we are familiar with. Nonetheless, the way we optimize is vastly different, considering policy distances rather than parameter distances. By ensuring policies do not drift too far when updating weights, we can perform more stable and consistent updates. 

Natural policy gradients come with a series of numerical challenges, particularly when dealing with large-scale optimizations (e.g., neural networks with large numbers of parameters). Also, a substantial number of approximations and simplifications are made in the theoretical foundation; practice may be more unruly. For real-world implementations, Proximal Policy Optimization is typically preferred nowadays.

Nonetheless, an understanding of natural gradients is fundamental for those wishing to understand the state-of-the-art in Reinforcement Learning.

\section{Further reading}
For the origins of Natural Policy gradients, I would suggest reading the foundational papers by Amari (1998) and Kokade (2001), as well as the more recent reflection by Martens (2020).

\begin{itemize}
\item Amari, S. I. (1998). Natural gradient works efficiently in learning. Neural computation, 10(2), 251–276. 
\item Kakade, S. M. (2001). A natural policy gradient. Advances in neural information processing systems, 14.
\item Martens, J. (2020). New insights and perspectives on the natural gradient method. The Journal of Machine Learning Research, 21(1), 5776–5851.
\end{itemize}

\noindent  In terms of lecture slides, I found the following ones particularly helpful.

\begin{itemize}
\item Levine, S. Advanced Policy Gradients (CS 285). UC Berkeley.
\item Achiam, J. (2017). Advanced Policy Gradient Methods. UC Berkeley.
\item Fragkiadaki, K. Natural Policy Gradients (CMU 10–403). Carnegie Mellon.
\end{itemize}

\noindent Finally, the following posts provide great explanations from different angles.

\begin{itemize}
\item Kristia, A. (2018). Natural Gradient Descent. URL: \url{https://agustinus.kristia.de/techblog/2018/03/14/natural-gradient/}
\item Vitay, J. Natural Gradients. URL: \url{https://julien-vitay.net/deeprl/NaturalGradient.html}
\item Jan Peters (2010). Policy Gradient Methods. Scholarpedia, 5(11):3698. URL: \url{http://www.scholarpedia.org/article/Policy_gradient_methods}
\item OpenAI (2018). Trust Region Policy Optimization. URL: \url{https://spinningup.openai.com/en/latest/algorithms/trpo.html#id2}
\end{itemize}

 \end{document}